%% file: main_archive.tex
\documentclass[10pt,twocolumn,letterpaper]{article}

\usepackage{iccv}
\usepackage{times}
\usepackage{epsfig}
\usepackage{graphicx}
\usepackage{amsmath}
\usepackage{amssymb}
\usepackage{booktabs}
\usepackage{algorithm}
\usepackage{algorithmicx,algpseudocode}
\usepackage{array,multirow}
\usepackage[table]{xcolor}
\usepackage[pagebackref=true,breaklinks=true,colorlinks,bookmarks=false]{hyperref}
\input{math_commands}

\usepackage[capitalize]{cleveref}
\iccvfinalcopy % *** Uncomment this line for the final submission

\ificcvfinal\fi

\begin{document}

%%%%%%%%% TITLE
\title{Efficient Self-supervised Continual Learning with Progressive Task-correlated Layer Freezing}

\author{Li Yang$^1$, Sen Lin$^2$, Fan Zhang$^1$, Junshan Zhang$^3$, and Deliang Fan$^1$\\
$^1$School of Electrical, Computer and Energy Engineering, Arizona State University\\
$^2$Department of ECE,
The Ohio State University \\
$^3$Department of ECE, University of California, Davis \\
{\tt\small  lyang166@asu.edu, 
lin.4282@osu.edu,
fzhang95@asu.edu,
jazh@ucdavis.edu,
dfan@asu.edu}}
% {\tt\small  lyang166@asu.edu, asrakin@asu.edu, dfan@asu.edu}}
% \and

\maketitle
\ificcvfinal\thispagestyle{empty}\fi

%%%%%%%%% ABSTRACT
\begin{abstract}
Inspired by the success of Self-supervised learning (SSL)  in learning visual representations from unlabeled data,  
a few recent works have studied SSL in the context of continual learning (CL), where multiple tasks are learned sequentially, giving rise to a new paradigm, namely self-supervised continual learning (SSCL). It has been shown that the SSCL outperforms supervised continual learning (SCL) as the learned representations are more informative and robust to catastrophic forgetting. However, if not designed intelligently, the training complexity of SSCL may be prohibitively high due to the inherent  training cost of SSL. In this work, by investigating the task correlations in SSCL setup first, we discover an interesting phenomenon that, with the SSL-learned background model, the intermediate features are highly correlated between tasks. Based on this new finding, we propose a new SSCL method with layer-wise freezing which progressively freezes partial layers with the highest correlation ratios for each task to improve training computation efficiency and memory efficiency. Extensive experiments across multiple datasets are performed, where our proposed method shows superior performance against the SoTA SSCL methods
under various SSL frameworks. For example, compared to LUMP, our method achieves 12\%/14\%/12\% GPU training time reduction, 23\%/26\%/24\% memory reduction, 35\%/34\%/33\% backward FLOPs reduction, and 1.31\%/1.98\%/1.21\% forgetting reduction without accuracy degradation on three datasets, respectively.  
\end{abstract}

%%%%%%%%% BODY TEXT
\section{Introduction}
\label{sec:intro}
Self-supervised learning (SSL) has achieved great success for unsupervised visual representation learning, which aims to learn representation  without the need for human annotations.
Recent studies (e.g., \cite{chen2020simple,chen2021exploring,he2020momentum,zbontar2021barlow}) have shown that 
SSL can achieve comparable or even better performance than the supervised learning counterpart, by utilizing different augmentation views from the same images to generate and optimize contrastiveness. However, SSL still suffers from two major challenges: (1) SSL typically assumes that all training data is available during the training process and learns offline with large amounts of data and resources.  In order to integrate new knowledge into the model, the current SSL methods  need to train repeatedly on the entire dataset. This may hinder their applications in some real-world scenarios, where new unlabeled data is made available progressively over time and old data becomes unavailable. Also, the learners must be able to cope with non-stationary data in a continuous manner when they are exposed to tasks with varying distributions of data. (2) Compared to supervised learning, obtaining the trained model with the same performance requires much larger training cost in various settings (e.g., heavy model size, larger batch-size, longer training epochs, etc,.).  

Fortunately, Continual Learning (CL)~\cite{kirkpatrick2017overcoming} provides a promising solution to address the first challenge. Notably, CL aims to incrementally update a model  over a sequence of tasks, performing knowledge transfer from the old tasks to the new ones without catastrophic forgetting. A large body of works has been proposed (e.g.,~\cite{rusu2016progressive, zenke2017continual,yoon2017lifelong,aljundi2019gradient,buzzega2020dark,zeng2019continual,farajtabar2020orthogonal,saha2021gradient,lin2022trgp,lin2022beyond}) for the supervised continual learning (SCL) paradigm. Most recently, a few works~\cite{fini2022self,gomez2022continually,madaan2021representational,hu2022well} have emerged to study CL for the self-supervised learning paradigm, named as \textit{Self-Supervised Continual Learning} (SSCL), which demonstrates that self-supervised visual representations are more robust to catastrophic forgetting compared to supervised learning. Specifically, CaSSLe~\cite{fini2022self} and PFR~\cite{gomez2022continually} propose a similar method that designs a temporal projection module to ensure that the newly learned feature space preserves the information of the previous one. UCL~\cite{madaan2021representational} adopts the Mixup~\cite{zhang2017mixup} technique to interpolate data between the current task and previous tasks’ instances to alleviate catastrophic forgetting of the learned representations. However, these works directly combine the existing SSL frameworks with supervised continual learning techniques (e.g, knowledge distillation, mixup, memory replay, etc,.). Inheriting from the first challenge of SSL as mentioned above, these works still suffer from large training cost and the catastrophic forgetting issue still remains.

In this work, we seek to tackle this challenge and  reduce training costs while mitigating catastrophic forgetting for SSCL. Towards this end, by first analyzing the task correlations based on gradient projection in SSCL, we find that the intermediate representations learned by self-supervised learning are \textit{highly correlated and varied} among tasks in comparison to SCL. Inspired by this new finding, we propose a new SSCL method with \textbf{progressive task-correlated layer freezing (PTLF)} during training for each task to reduce training time and memory cost. Specifically, we first define the \textit{task correlation ratio} according to the gradient projection norm to formally characterize the correlation between the current task and prior tasks. Then, the top-ranked layers with higher task correlation ratios among tasks are progressively frozen during the self-supervised continual learning process for each task. 

In the experiments, we validate the proposed method against the state-of-the-art SSCL methods on mutiple benchmarks, including Split CIFAR-10, Split CIFAR-100, Split TinyImageNet and ImageNet-100. The experimental results on various SSL frametworks have shwon that our method demonstrates superior performance of training efficiency and forgetting over the baseline methods, with comparable or even better accuracy. For example, in comparison to LUMP~\cite{madaan2021representational}, by using SimSiam~\cite{chen2021exploring} SSL framework, our method achieves 12\%/14\%/12\% GPU training time reduction, 23\%/26\%/24\% memory reduction, 35\%/34\%/33\% backward FLOPs reduction, and 1.31\%/1.98\%/1.21\% forgetting reduction without accuracy degradation on three datasets, respectively.

\section{Related Work}
\label{sec:rel_work}

\subsection{Self-supervised learning}
\label{sec:ssl}
Self-supervised learning  aims to learn visual representation without data labeling cost. Recent advances~\cite{he2020momentum,chen2020simple,grill2020bootstrap,zbontar2021barlow,chen2021exploring} show that self-supervised learning can achieve similar or even better performance than supervised representation learning.
A common strategy of these methods is to learn representations that are invariant under different data augmentations by maximizing their similarity with contrastive loss optimization. However, 
these approaches require large-sized batches and negative samples.
%, which  limits their applications in the continual learning domain. 
SimSiam~\cite{chen2021exploring} addresses this issue by utilizing the stop-gradient technique to prevent the collapsing of Siamese networks. The Siamese network consists of an encoder network $f$ and a prediction MLP $h$, where the encoder includes a backbone model (e.g., ResNet~\cite{he2016deep}) and a projection MLP. Given two randomly augmented views of $x_1$ and $x_2$ from an input image $x$, Simsiam aims to minimize the negative cosine similarity between the predictor output $p_1$ ($p_1 = f(h(x_1)) $) and the projector output $z_2$ ($z_2 = f(x_2)$) with a symmetrized loss as: 
\begin{equation}
\small	
    L_{SSL} = \frac{1}{2}D(p_1, \mbox{stopgrad}(z_2)) + \frac{1}{2}D(p_2, \mbox{stopgrad}(z_1))
\label{eqt:simsiam}
\end{equation}
where $D$ is a negative cosine similarity function. 
Given the distorted versions of an instance, BarlowTwin~\cite{zbontar2021barlow} minimizes the redundancy between their embedding vector components while conserving the maximum information. This can be achieved by making the cross-correlation matrix, computed between the outputs of two identical networks, closer to the identity matrix, through the minimization of the following loss: 
\begin{equation}
    L_{SSL} = \sum_{i}(1-C_{ii})^{2} + \lambda \sum_{i}\sum_{j\neq i}C_{ij}.
\label{eqt:barlowtwin}
\end{equation}
Here $\lambda$ is a positive constant scaling factor and $C$ is the cross-correlation matrix computed between the outputs of the two identical networks along the batch dimension.
Since SimSiam and BarlowTwim
have no requirements for large batch size and negative samples, in this work, we adopt these two works as base learning methods for self-supervised continual learning.

\begin{figure*}[t]
    \centering
\includegraphics[width=0.86\linewidth]{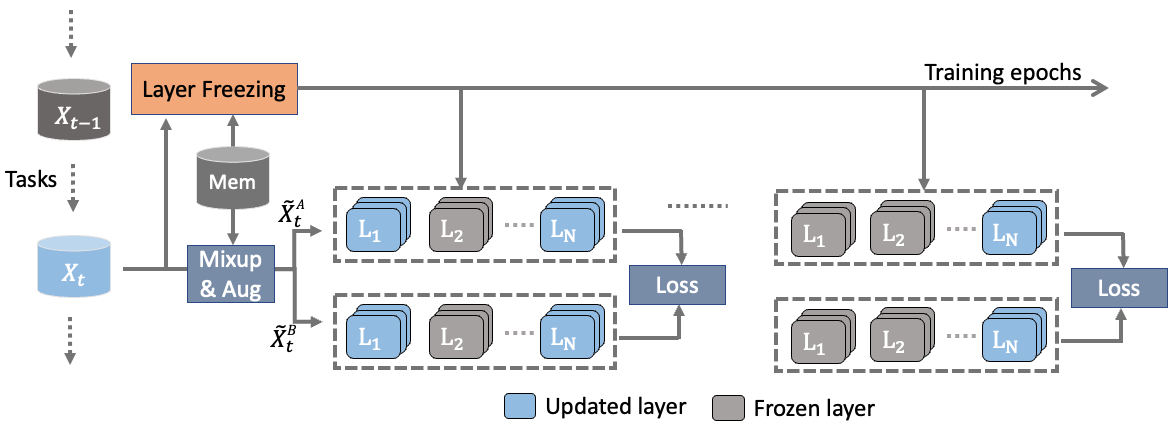}   
% (a) The prior work LUMP~\cite{madaan2021representational} 
% \includegraphics[width=0.42\linewidth]{figures/overview_new.pdf}
    \caption{The overview of our proposed method which progressively freezes partial layers during the whole training process for each task.} 
\label{fig:overview}
\end{figure*}

\subsection{Continual Learning}
Plentiful continual learning methods have been developed in supervised learning and can be generally divided into three categories: 1) \emph{Regularization-based methods} (e.g., \cite{aljundi2018memory,lee2017overcoming, kirkpatrick2017overcoming}) preserve the knowledge of old tasks by adding an additional regularization term in the loss function, in order  to constrain the weight update when learning the new task. For example, Elastic Weight Consolidation (EWC) \cite{kirkpatrick2017overcoming} regularizes the update on the important weights that are evaluated using the Fisher Information matrix. 2) \emph{Structure-based methods} (e.g., \cite{yoon2017lifelong, yoon2020scalable, serra2018overcoming, yang2021grown}) adapt different model parameters or architectures with a sequence of tasks. 
3) \emph{Memory-based methods} can be further divided into memory-replay methods and orthogonal-projection based methods. Memory-replay methods (e.g., \cite{riemer2018learning, guo2020improved,chaudhry2018efficient}) store and replay the old tasks data when learning the new task, while
orthogonal-projection based methods
(e.g., \cite{zeng2019continual,farajtabar2020orthogonal,saha2021gradient,lin2022trgp,lin2022beyond}) update the model for each new task in the orthogonal direction to the  subspace spanned by inputs of old tasks.
%, by storing and projecting the gradient or subspace of old tasks on the current task. 

More recently, a few works~\cite{rao2019continual,fini2022self,gomez2022continually,madaan2021representational,hu2022well} have  emerged to tackle the problem of self-supervised continual learning.
They show that self-supervised continual learning can mitigate catastrophic forgetting and learn more general representations compared to supervised continual learning. Specifically, 
\cite{rao2019continual} learned task-specific representations on shared parameters. 
However, it is restricted to simple low-resolution tasks and not scalable to standard CL benchmark datasets. In addition,  CaSSLe~\cite{fini2022self} and PFR~\cite{gomez2022continually} propose a similar method that designs a temporal projection module to ensure that the newly
learned feature space preserves the information of the previous one. LUMP~\cite{madaan2021representational} adapts the Mixup~\cite{zhang2017mixup} technique to interpolate data between the current task and previous tasks’ instances to alleviate catastrophic forgetting for unsupervised representations:
\begin{equation}
    \tilde{x}_{t,i} = \lambda \cdot x_{t,i} + (1-\lambda) \cdot x_{M,l},
\end{equation}
where $x_{M, l}$ denotes the old task data selected using uniform sampling from replay buffer $M$ and $\lambda$ is randomly sampled from a $Beta$ distribution.
However, these works directly combine
the existing self-supervised with continual learning techniques (e.g, knowledge distillation, mixup, memory replay,
etc,.) that still suffer from large training costs, and the forgetting issue remains as well. In this work, we aim to tackle these 
concerns.

\subsection{Layer Freezing}
There existed several works on accelerating the training of deep neural networks for one single task by using layer freezing techniques ~\cite{liu2021autofreeze,he2021pipetransformer,wang2022efficient,aguilar2020knowledge,yuan2022layer}. These works are motivated by the fact that front layers mainly extract general features of the
raw data (e.g., the shape of objects) and are easier to well-trained, while deeper layers are more task-specific and capture complicated features
output from front layers. Specifically, 
Liu et al.~\cite{liu2021autofreeze} propose to automatically freeze layers during training according to the parameter gradients. Wang et al.~\cite{wang2022efficient} adopt knowledge distillation~\cite{aguilar2020knowledge} to guide the layer freezing schedule. Yuan et al.~\cite{yuan2022layer} apply layer freezing on sparse training. However, these works only focus on one single task. It is a common setting that progressively freezes layers in descending layer index order. Different from all these related works, we show that layer freezing in SSCL needs to consider task correlations.

\section{Method}
\label{sec:method}
\subsection{Problem Setup}
In supervised continual learning, a model continuously learns from a sequential data stream in which new tasks (namely, classification tasks with new classes) are added over time. More formally, consider a sequence of tasks $\{1, 2, ..., T\}$ where the task at time  $t$ comes with training data $D_t = \{\vx_{t,i}, \vy_{t,i}\}_{i=1}^{N_t}$. 
Note that each task $t$ can contain a sequence of classes. 
We denote $f(\cdot)$ as the operation of a feature extractor and $h(\cdot)$ as a classifier model. 
%The model gets to observe tasks from $t_i$ to $t_N$ sequentially.
The main objective is to optimize the parameter $\vw$ of both the feature extractor and the classifier: 
\begin{equation}
    \mbox{min}_{\vw_f,\vw_h} ~ \sum_{t=1}^T\sum_{i = 1}^{N_t}\gL_t(h(f(\vx_{t,i})),\vy_{t,i})
\end{equation}
where $\gL_t(\cdot)$ is cross entropy loss function in general.

In contrast, self-supervised continual learning does not require data labels during training. The objective is to learn a general representation that is invariant to augmentations on all tasks, which can be formulated as:
\begin{equation}
    \mbox{min}_{\vw_f} ~ \sum_{t=1}^T\sum_{i = 1}^{N_t}\gL_t(f(\vx_{t,i}^{1},\vx_{t,i}^{2}))
\end{equation}
where $\vx_{t,i}^{1}$ and $\vx_{t,i}^{2}$ are augmented images generated from $\vx_{t,i}$. The choices of the feature extractor and loss function depend on the self-supervised learning method, such as SimSiam and BarlowTwin as shown in \cref{eqt:simsiam} and \cref{eqt:barlowtwin}, respectively. After the feature extractor is learned on all tasks, following \cite{madaan2021representational}, we use K-nearest neighbor (KNN) classifier~\cite{wu2018unsupervised} or linear classification to evaluate the performance of the learned representation.
In this work, we investigate self-supervised continual learning on the setting of \textbf{class-incremental learning}, where the model owns a single classifier and task identifiers are not provided during inference.

\begin{figure*}
    \centering
    \includegraphics[width=0.24\linewidth]{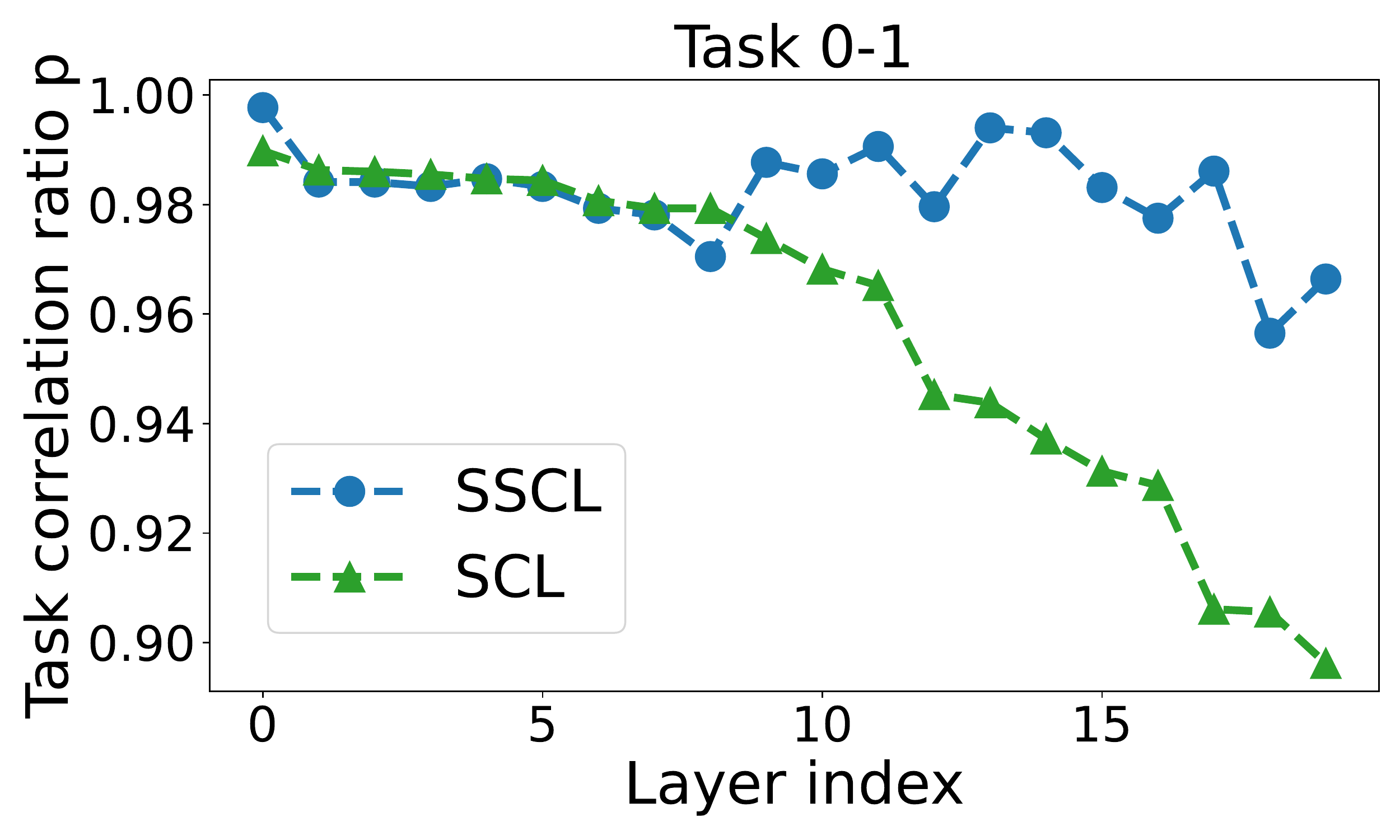} 
    \includegraphics[width=0.24\linewidth]{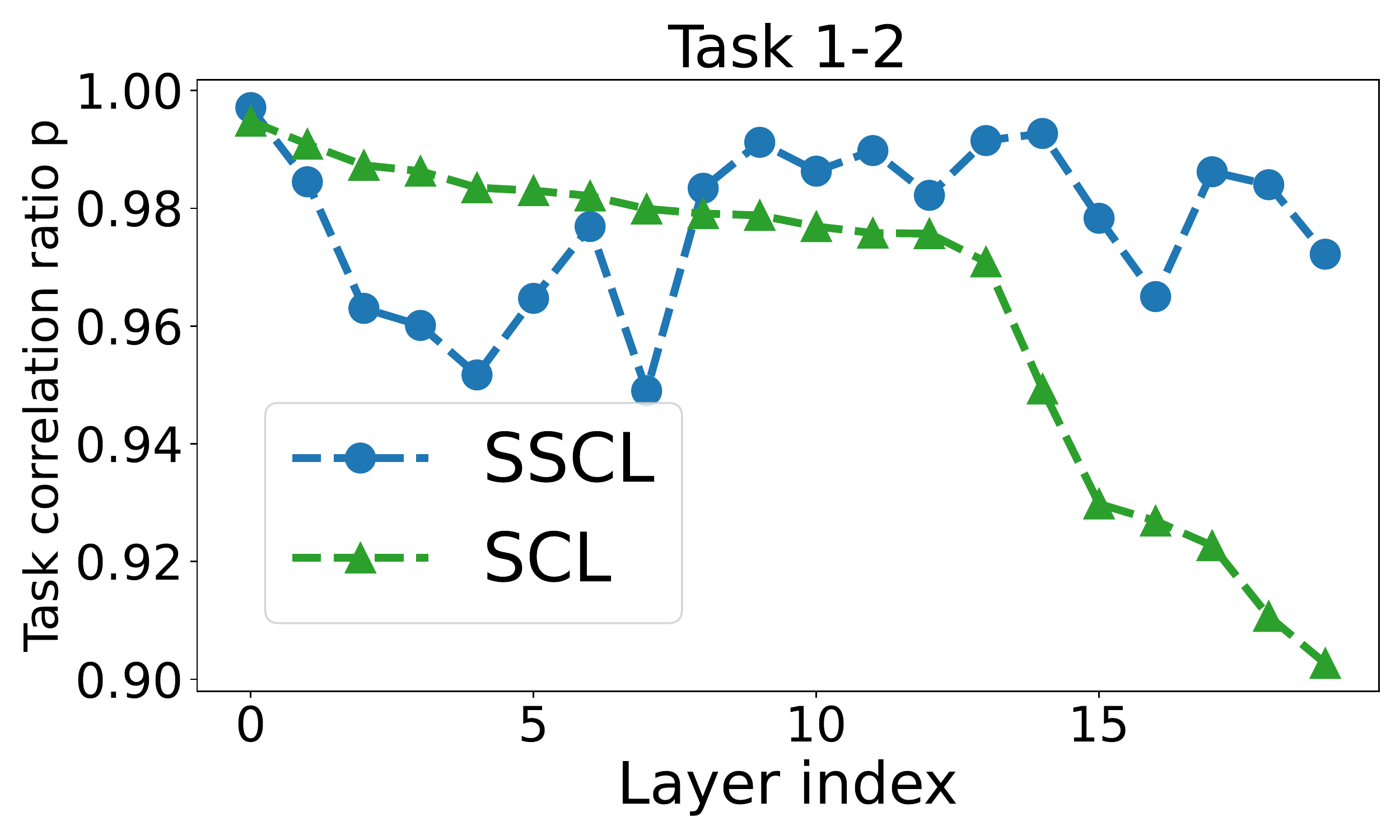} 
    \includegraphics[width=0.24\linewidth]{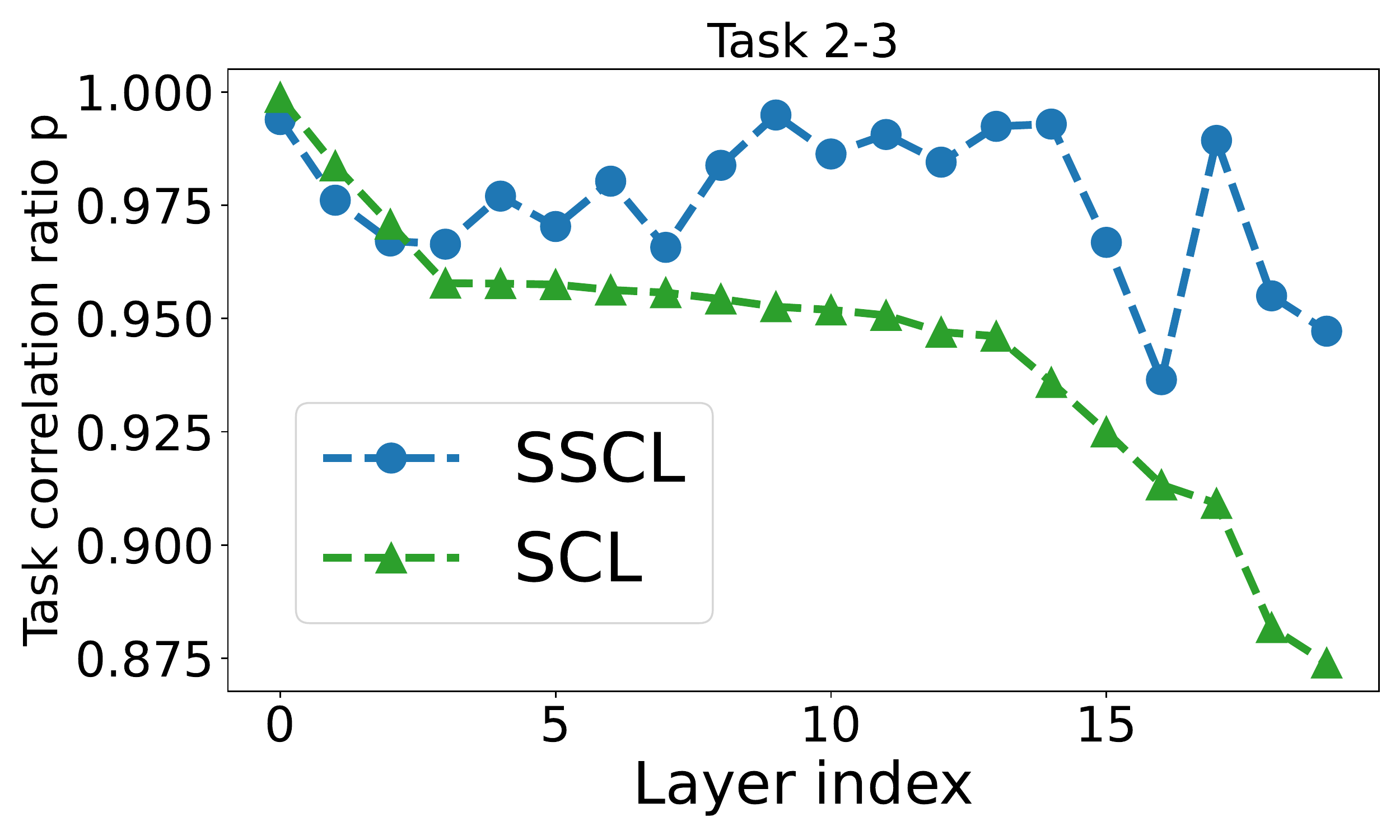}
    \includegraphics[width=0.24\linewidth]{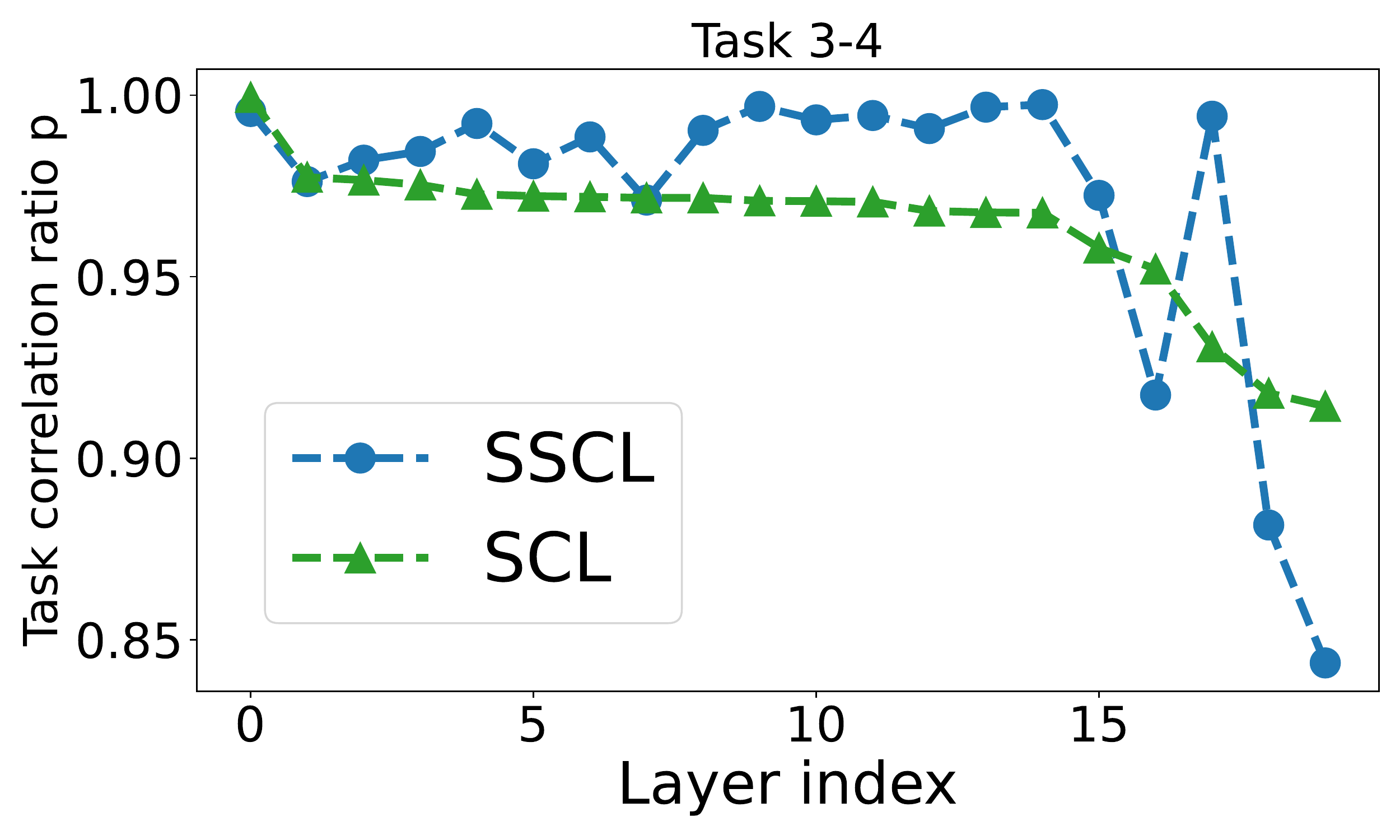}
    \caption{Layer-wise correlation ratio between two tasks on Split CIFAR-10 (5 tasks setup) using ResNet18 for SSCL and SCL respectively. } 
    % \textcolor{red}{can we make the legend and `task x-y' of all plots to be much larger and using different symbols to easily distinguish even in B/W?}}
\label{fig:corr}
\end{figure*}
\subsection{Proposed Method}

\subsubsection{Overview}
For the first task, following the general setting~\cite{rao2019continual,fini2022self,gomez2022continually,madaan2021representational}, we train the model as the self-supervised learning method without modification. Then for the remaining tasks that are learned sequentially, as shown in the \cref{fig:overview}, we adopt the memory replay-based method~\cite{buzzega2020dark} to uniformly store data of each task with a fixed buffer size (i.e., 256) and the mixup technique from LUMP~\cite{madaan2021representational}. Furthermore,
to improve the training computation and memory efficiency while mitigating catastrophic forgetting, we propose to \textit{progressively freeze} layers during the whole training process for each task. Here the frozen layers are determined by \textit{task correlation}. 
The detailed method will be illustrated below.
% will not be updated during backward for the current task, which achieves actual training computation and time reduction in GPU.

\subsubsection{Layer freezing via task correlation}

% {Progressive Task-correlated Layer Freezing (PTLF)}}

It is well-known that the training cost of SSL models is notoriously high, and this issue is further exacerbated in SSCL where new tasks arrive continuously. How to improve the training efficiency of SSCL becomes a critical question that needs to be solved, in order to facilitate the development and application of SSCL methods in practice. Towards this end, prior works~\cite{rao2019continual,fini2022self,gomez2022continually,madaan2021representational} have revealed that the learned representations of SSCL are more general and robust to catastrophic forgetting than the representations of SCL. We conjecture the reason behind is the learned intermediate features for each layer between the current task and prior tasks are highly correlated with each other. Accordingly, if the new task $t$ has strong similarity with old tasks in some layers, it is possible that not updating these layers will not significantly affect the learning performance of this new task. Thus inspired, we raise the question: \textit{Can we leverage the generality of the learned representations from SSL and freeze the highly correlated layers during training for each task to improve the training efficiency?}

To answer this question, motivated by prior gradient orthogonal-projection based methods~\cite{lin2022trgp,lin2022beyond} on SCL, we first investigate the correlation of tasks according to gradient projection. 
Specifically, to formally characterize the correlation between the current task and prior tasks, we  \textbf{define the task correlation ratio} in layer-wise as:

\begin{equation}
   r_l = \frac{\|\proj_{S_t^l}(\nabla \gL_t(\vw^l_{t-1}))\|_2}{\|\nabla\gL_t(\vw^l_{t-1}))\|_2} 
\label{eqt:ratio}
\end{equation}
where $\proj_{S_t^l}$ denotes the projection on the input subspace $S_t^l$ of prior tasks ($ 1,2,...,t-1$) on $l^{th}$ layer, and $\vw^l_{t-1}$ represents the $l^{th}$ layer weight in the model before learning task $t$. Here $\proj_{S}(A)=AB(B)'$ for some matrix $A$ and $B$ is the bases for $S$. Due to the fact that the gradient lies in the span of the input~\cite{saha2021gradient}, if the task correlation ratio $r_l \in (0,1)$ has a large value, it implies that the current task $t$ and prior tasks may have sufficient common bases in $l^{th}$ layer between their input subspaces and hence are strongly correlated. To quantitatively evaluate the task correlation on SSCL, we conduct the experiments on three settings (i.e., Split CIFAR-10, Split CIFAR-100, Split TinyImageNet) by using prior representative work LUMP~\cite{madaan2021representational}. As shown in \cref{fig:corr}, we observe that: 

\textbf{Observations:} \textit{
1) the variance of the task correlation ratio in SSCL is smaller than the counterparts in SCL; 2) the correlation ratios of SSCL are larger than the counterparts in SCL for most layers; 3) the correlation ratios of SCL consistently follow an ascending order, while the counterparts in SSCL are more varied that are usually higher for top and middle layers. } 

The first two observations help to further explain that the learned representations of SSCL are more general than SCL. Moreover, the third observation indicates that following ascending order to freeze layer in supervised learning is a good choice~\cite{brock2017freezeout,liu2021autofreeze,yuan2022layer} since the correlation ratios of the top layers are always larger than the later ones. However, for the SSCL, layer freezing needs to consider task correlations between tasks. In the experiments, we also evaluate that task-correlated layer freezing could show better accuracy than conventional layer freezing in ascending order in \cref{sec:abla}.

\paragraph{Subspace construction via memory replay data.}

Prior orthogonal-projection based methods (e.g., \cite{saha2021gradient,lin2022trgp}) target on task-incremental learning, which calculate and then store the bases of the input subspace of each prior task individually for orthogonal gradient descent. Such subspace storage consumes large memory costs, especially for the large models of self-supervised learning. For example, ResNet18  requires 116MB on CIFAR-10 dataset. Benefiting from utilizing data replay mechanism, we can construct the subspace of prior tasks on-the-fly instead of storing the corresponding bases. In practice, before training the current task, we calculate the bases of the subspace for the data in the replay buffer using Singular Value Decomposition (SVD) on the representations. Specifically, given the model $\vw_{t-1}$ before learning task t, we construct a representation matrix $\mR_t^l=[\vx_{r,1}^l,...,\vx_{r,n}^l]\in\sR^{m\times n}$ with $n$ samples from memory replay buffer, where each $\vx_{r,i}^l\in\sR^m$, is the representation at layer $l$ by forwarding the sample $\vx_{r,i}$ through the network. Then, we apply SVD to the matrix $\mR_t^l$, i.e., $\mR_t^l=\mU_t^l \mSigma_1^l (\mV_t^l)'$,
where $\mU_t^l=[\vu_{t,1}^l,...,\vu_{t,m}^l]\in \sR^{m\times m}$ is an orthogonal matrix with left singular vector $\vu_{t,i}^l\in \sR^m$,  $\mV_t^l=[\vv_{t,1}^l,...,\vv_{t,n}^l]\in \sR^{n\times n}$ is an orthogonal matrix with right singular vector $\vv_{t,i}^l\in \sR^n$, and $\mSigma_t^l\in\mR^{m\times n}$ is a rectangular diagonal matrix with non-negative singular values $\{\sigma^l_{t,i}\}_{i=1}^{\min\{m,n\}}$ on the diagonal in a descending order. To obtain the bases for subspace $S_t^l$, we use $k_t^l$-rank matrix approximation to pick the first $k_t^l$ left singular vectors in $\mU_t^l$, such that the following condition is satisfied for a threshold $\eta_{th}^l\in (0,1)$:
\begin{align}
    \|(\mR_t^l)_{k_t^l}\|^2_F\geq \epsilon_{th}^l\|\mR_t^l\|^2_F
\end{align}
where $(\mR_t^l)_{k_t^l}=\sum_{i=1}^{k_t^l} \sigma^l_{t,i}\vu_{t,i}^l (\vv^l_{t,i})'$ is a $k_t^l$-rank ($k_t^l\leq r$) approximation of the representation matrix $\mR_t^l$ with rank $r\leq \min\{m,n\}$, and $\|\cdot\|_F$ is the Frobenius norm. Then the bases  for subspace $S_t^l$ can be constructed as $\mB_t^l=[\vu_{t,1}^l,...,\vu_{t,k_t^l}^l]$.

\subsubsection{Progressive task-correlated freezing} Based on the proposed task-correlation metric, we further propose \textbf{progressive task-correlated freezing} in SSCL to progressively  freeze partial layers with the highest correlation ratios during training for each task, in order to enhance the training computation and memory efficiency. Specifically, define the initial freeze ratio as $k_i$ and final freeze ratio as $k_f$, which denote the ratio of the number of frozen layers to the number of layers in the neural network. The total number of training epochs is $N$ and the current training epoch is $n$. We adopt cosine annealing to progressively increase the freeze ratio in epoch-wise:
\begin{equation}
    k_{n} = k_i + \frac{1}{2}(k_f - k_i)(1+ cos(\frac{n}{N}\pi))
\label{eqt:cos_annel}
\end{equation}
% \textcolor{red}{($k_{curr}$ or $k_n$)}
where $k_{n}$ is the freeze ratio for the current epoch. In our experiments, we set the initial and final freeze ratios as 0 and 0.4 for all tasks by default.

Following that, once getting the freeze ratio for the current epoch, we adopt the following strategies to progressively and accumulately freeze the layers: 1) the layers with the highest task-correlation ratio under the current freeze ratio $k_{n}$ will be frozen; 2) the frozen layers of prior epochs will be unchanged, and we will gradually increase the number of frozen layers according to the freeze ratio difference $(k_n - k_{n-1})$. This can be achieved by using a TopK function according to the layer-wise task correlation to select the layers to freeze: 
\begin{equation}
    F_{n} = \{l| r_{n}^l \in \mbox{TopK}(R_{n}, k_n - k_{n-1})\}
\label{eqt:top-k}
\end{equation}
where $R_{n}$ denotes a set of task correlations across all unfrozen layers in current epoch $n$ and $r_{n}^l$ is the task-correlation ratio for $l^{th}$ layer as defined in \cref{eqt:ratio}. By doing so, we could generate a set of indexes of new frozen layers $F_{n}$ for each task.
Importantly, 
% it is worth noting that
% that the proposed layer-freezing method can be used in various freeze pattern (e.g., layer-wise, channel-wise, element-wise). 
One practical reason that we choose layer-wise freezing is that layer-wise freezing could enable actual training speedup in GPU by using general deep learning frameworks (e.g, Tensorflow, Pytorch). In addition, we find that if applying the proposed layer-wise weight freezing in supervised continual learning (SCL) setup, it will cause clear accuracy degradation, which also advocates that the representation learned by SSCL is more general and robust. The detailed analysis can be found in \cref{sec:granul}.

\begin{table}[t]
\centering
\caption{One-shot updating \textit{vs} per-epoch updating.}
\scalebox{0.88}{
\begin{tabular}{ccccc}
\toprule
\multirow{2}{*}{Setting} & \multicolumn{2}{c}{Split CIFAR-10} & \multicolumn{2}{c}{Split CIFAR-100} \\ \cmidrule{2-5} 
                   & Accuracy & Forgetting & Accuracy & Forgetting \\ \cmidrule{2-5} 
One-shot   &     89.73     &      0.86     &     80.54     &  2.24         \\
Per-epoch &      89.81    &      0.80     &      80.33    &      2.10     \\ \bottomrule
\end{tabular}}
\label{tab:one-shot}
\end{table}

\begin{figure}[hbt!]
    \centering
\includegraphics[width=0.7\linewidth]{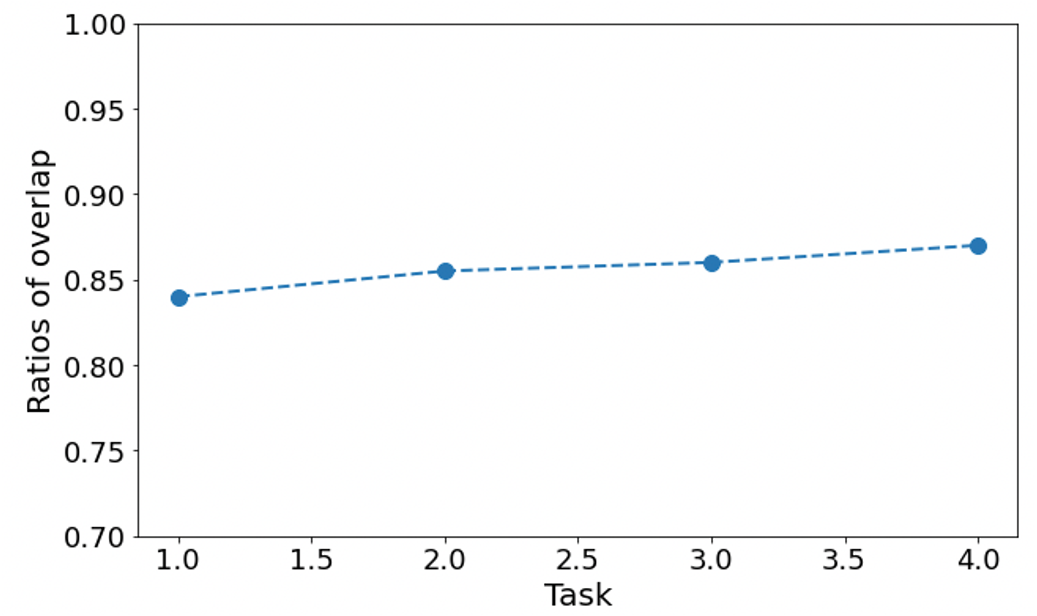}
    \caption{The overlap of layer selection for different tasks}
\label{fig:overlap}

\end{figure}

\paragraph{Layer-wise weight freezing in one-shot.}
Considering the training procedure for each task, there are mainly two choices to achieve layer-wise weight freezing: 1) \textit{one-shot updating}, which only makes the frozen decision before training and fixes the selection during the entire training procedure; 2) \textit{per epoch updating}, which updates the layer freezing decision for the unfrozen layers so far in each epoch during training. To understand the learning behaviors herein, we conduct experiments for both cases on Split CIFAR-10 and Split CIFAR-100 settings, respectively, by using ResNet-18 as the backbone model. As shown in \cref{tab:one-shot}, we find that both cases have similar performance. Moreover, as shown in \cref{fig:overlap}, it is interesting to see that the average overlap ratios of the layer selection among all the epochs for all tasks are very high (i.e., larger than 0.8), indicating that the layer selection decision during the learning process is stable. Based on these observations, we adopt the \textit{one-shot} setting where we only conduct the layer selection one time for each task, such that the extra computation cost for the sub-space calculation is minimal.

In a nutshell, the details of the proposed method are summarized in \cref{tab:algo}.

\begin{algorithm}[tb]
\footnotesize
	\caption{Efficient Self-supervised Continual Learning with Progressive Task-correlated Layer Freezing}
	\label{alg1}
 	\begin{algorithmic}[1]
 	\State Input: task sequence $\sT=\{t\}_{t=1}^T$;
 	\State Learn the first task using vanilla stochastic gradient descent;
 	% \State Extract the bases $\{B_l^1\}$ based on SVD using the learnt model $\vw^1$;
 	\For{each task $t \in \{1, T\}$}
 	    \State Extract the bases $\{B_t^l\}$ based on SVD using the learnt model $\vw^{t-1}$; 
 	    \State Generate the layer-wise task correlation ratio by solving \cref{eqt:ratio};
 	    \For{epoch=1, 2,...}
                \State Gradually increase the freeze ratio via cosine annealing based on \cref{eqt:cos_annel};
                \If{layer $l$ is not frozen}
                    \State Update the parameters of the layer  
                \EndIf
 	    \EndFor
        \EndFor
	\end{algorithmic}
\label{tab:algo}
\end{algorithm}

% \linote{First describe layer-wise frozen idea.
% \begin{itemize}
%     \item Why we do weight frozen. Before training or during training. 
%     \item Why we do layer-wise
%     \item In the class-incremental learning, we use memory replay to calculate the projection with minimum extra computation cost. By doing so, the memory cost is largely reduced compared to the original orthogonal gradient descent method.
% \end{itemize}
% Need to highlight our finding that such method is uniqueness to self-supervised learning. }

% \linote{second.for the rest layers, update weights with regularization constraint}
\begin{table*}[ht]
\scriptsize
\centering
\caption{Accuracy and forgetting of the learned representations on Split CIFAR-10, Split CIFAR-100 and Split Tiny-ImageNet on ResNet-18 architecture with KNN classifier. All the values are measured by computing mean and standard deviation across three trials. Note that, we use the layer freezing ratio as 0.4 by default for all our results. }
\scalebox{1.0}{
\begin{tabular}{lcllllll}
\toprule
\multicolumn{2}{c}{Method} &
  \multicolumn{2}{c}{SPLIT CIFAR-10} &
  \multicolumn{2}{c}{SPLIT CIFAR-100} &
  \multicolumn{2}{c}{SPLIT TINY-IMAGENET} \\ \midrule
 &
   &
  \multicolumn{1}{c}{Accuracy} &
  Forgetting &
  \multicolumn{1}{c}{Accuracy} &
  Forgetting &
  \multicolumn{1}{c}{Accuracy} &
  Forgetting 
\\ \midrule
\multirow{8}{*}{\rotatebox[origin=c]{90}{Simsiam}} & Finetune  & 90.11 ( $\pm 0.12$) & 5.43 ($\pm 0.08$) & 75.42 ($\pm 0.78$) & 10.19 ($\pm 0.78$)  & 71.07 ($\pm 0.20$) & 9.48 ($\pm 0.56$)  \\\cmidrule{2-8}
\multicolumn{1}{c}{}                         & PNN~\cite{rusu2016progressive}       &  90.93 ($\pm 0.22$) & - & 66.58 ($\pm 1.0$) & -  & 62.15 ($\pm 1.35$) & - \\
\multicolumn{1}{c}{}                         & SI~\cite{zenke2017continual}        &  92.75 ($\pm 0.06$) & 1.81 ($\pm 0.21$) & 80.08 ($\pm 1.3$)& 5.54 ($\pm 01.3$)  & 72.34 ($\pm 0.42$)  & 8.26 ($\pm 0.64$)   \\
\multicolumn{1}{c}{}                         & DER~\cite{buzzega2020dark}       &  91.22 ($\pm 0.3$) & 4.63 ($\pm 0.26$)  & 77.27 ($\pm 0.30$) & 9.31 ($\pm 0.09$)  & 71.90 ($\pm 1.44 $)  & 8.36 ($\pm 2.06 $) \\\cmidrule{2-8}
\multicolumn{1}{c}{}                         & CassLe~\cite{gomez2022continually}    & \textbf{91.04} ($\pm 0.24$) & 2.24 ($\pm 0.23$)  & 81.58 ($\pm 0.84$) & 5.02 ($\pm 1.12$)  & 75.77 ($\pm 1.74$) & 4.42 ($\pm 1.24$)  \\
\multicolumn{1}{c}{}                         & LUMP~\cite{madaan2021representational}      & 91.00 ($\pm 0.40$) & 2.92 ($\pm 0.53$)  & \textbf{82.30} 
($\pm 1.35$) & 4.71 ($\pm 1.52$) & 76.66 ($\pm 2.39$) & 3.54 ($\pm 1.04$)  \\
\multicolumn{1}{c}{}                         & Ours      & 91.03 ($\pm 0.44$) & \textbf{1.61} ($\pm 0.23$)  & 82.24 ($\pm 1.24$) & \textbf{2.73}($\pm 1.13$)  & \textbf{76.68} ($\pm 2.51$) &  \textbf{2.33}($\pm 1.14$) \\ \cmidrule{2-8}  
\multicolumn{1}{c}{}                         & Multitask &  95.76 ($\pm 0.08$) & - & 86.31 ($\pm 0.38$) & - &  82.89 ($\pm 0.49$) & - \\ \midrule
\multirow{8}{*}{\rotatebox[origin=c]{90}{BarlowTwin}}       & Finetune  & 87.72 ($\pm 0.32$) & 4.08 ($\pm 0.56$)  & 71.97 ($\pm 0.54$)  & 9.45 ($\pm 1.01$)  & 66.28 ($\pm 1.23$) & 8.89 ($\pm 0.66$)  \\\cmidrule{2-8}
 & PNN~\cite{rusu2016progressive}       & 87.52 ($\pm 0.33$) & -  & 57.93 ($\pm 2.98$)  & - & 48.70 ($\pm 2.59$) & -\\
 & SI~\cite{zenke2017continual}        & \textbf{90.21} ($\pm 0.08$)  & 2.03 ($\pm 0.22$)   & 75.04 ($\pm 0.63$) & 7.43 ($\pm 0.67$)  & 56.96 ($\pm 1.48$) & 17.04 ($\pm 0.89$)  \\
 & DER~\cite{buzzega2020dark}       & 88.67 ($\pm 0.30$) & 2.41 ($\pm 0.26$)  & 73.48 ($\pm 0.53$) & 7.98 ($\pm 0.29$)  & 68.56 ($\pm 1.47$) & 7.87 ($\pm 0.44$)  \\ \cmidrule{2-8}
 & Cassle~\cite{gomez2022continually}    & 89.04 ($\pm 0.34$) & 1.89 ($\pm 0.14$)  &  77.05 ($\pm 0.75$) & 2.47 ($\pm 0.44$)  & 71.76 ($\pm 0.65$) & 2.88 ($\pm 0.45$)  \\
 & LUMP~\cite{madaan2021representational}       & 89.72 ($\pm 0.30$) & 1.13 ($\pm 0.18$) & 80.24 ($\pm 1.04$) & 3.53 ($\pm 0.83$)  & 72.17 ($\pm 0.89$) & 2.43 ($\pm 1.00$)   \\
 & Ours      & 89.73 ($\pm 0.41$) & \textbf{0.92} ($\pm 0.23$)  & \textbf{80.54} ($\pm 0.88$) & \textbf{2.24} ($\pm 0.84$)  & \textbf{73.56} ($\pm 1.02$) &  \textbf{1.74} ($\pm 0.62$) \\ \cmidrule{2-8}
 & Multitask & 95.48 ($\pm 0.14$) & -  & 87.16 ($\pm 0.52$) & -  & 82.42 ($\pm 0.74$) & -  \\ \bottomrule
\end{tabular}}
\label{tab:main_results}
\end{table*}

\begin{table*}[h]
\scriptsize
\centering
\caption{Training time (measured time in NVIDIA A4000 GPU), memory cost, and computation FLOPs of the learned representations on Split CIFAR-10, Split CIFAR-100 and Split Tiny-ImageNet. }
\scalebox{1.0}{
\begin{tabular}{lclllllllll}
\toprule
\multicolumn{2}{c}{Method} &
  \multicolumn{3}{c}{SPLIT CIFAR-10} &
  \multicolumn{3}{c}{SPLIT CIFAR-100} &
  \multicolumn{3}{c}{SPLIT TINY-IMAGENET} \\ \midrule
 &
   &
  \multicolumn{1}{c}{Time} &
   Memory & FLOPs &
  \multicolumn{1}{c}{Time} &
   Memory & FLOPs &
  \multicolumn{1}{c}{ Time} &
   Memory & FLOPs \\ \midrule
\multirow{6}{*}{\rotatebox[origin=c]{90}{Simsiam}}    & PNN~\cite{rusu2016progressive}   & 1.35x & 1.35x  &  1.35x & 1.35x & 1.35x & 1.35x & 1.35x& 1.35x & 1.35x \\
& SI~\cite{zenke2017continual}       & 1.2x & 1.2x  &  1.2x & 1.2x & 1.2x & 1.2x  &  1.2x & 1.2x & 1.2x \\
& DER~\cite{buzzega2020dark}    & 1x & 1x  &  1x & 1x & 1x & 1x & 1x& 1x & 1x \\ 
& LUMP~\cite{madaan2021representational}    & 1x & 1x  &  1x & 1x & 1x & 1x & 1x& 1x & 1x \\
   & CassLe~\cite{gomez2022continually}       & 1.3x & 1.3x & 1.3x & 1.3x &  1.3x & 1.3x &  1.3 & 1.3x & 1.3x \\
    & Ours       & \textbf{0.88x} & \textbf{0.77x} & \textbf{0.68x} & \textbf{0.86x} & \textbf{0.74x} & \textbf{0.67x}& \textbf{0.88x} &  \textbf{0.76x} & \textbf{0.68x} \\ \midrule  
\multirow{6}{*}{\rotatebox[origin=c]{90}{\tiny{BarlowTwin}}}    
  & PNN~\cite{rusu2016progressive}   & 1.35x & 1.35x  &  1.35x & 1.35x & 1.35x & 1.35x & 1.35x& 1.35x & 1.35x \\
& SI~\cite{zenke2017continual}       & 1.2x & 1.2x  &  1.2x & 1.2x & 1.2x & 1.2x  &  1.2x & 1.2x & 1.2x \\
& DER~\cite{buzzega2020dark}    & 1x & 1x  &  1x & 1x & 1x & 1x & 1x& 1x & 1x \\ 
 & LUMP~\cite{madaan2021representational}    & 1x & 1x &  1x & 1x & 1x & 1x & 1x & 1x & 1x \\
 & Cassle~\cite{gomez2022continually}   & 1.3x &  1.3x &  1.3x & 1.3x &  1.3x &  1.3x & 1.3x &  1.3x &  1.3x  \\
 & Ours   & \textbf{0.87x} & \textbf{0.78x} & \textbf{0.65x} & \textbf{0.88x} & \textbf{0.79x} & \textbf{0.66x} & \textbf{0.88x}  & \textbf{0.75x} & \textbf{0.67x}  \\ \bottomrule 
\end{tabular}}
\label{tab:effic_results}
\end{table*}

\section{Experimental Results}
\label{sec:exp}
\subsection{Experimental setup}
\textbf{Baselines.} Prior SSCL works~\cite{rao2019continual,fini2022self,gomez2022continually,madaan2021representational} have shown that the SSCL methods outperform SCL counterparts in class incremental learning. Thus we compare with multiple self-supervised continual learning baselines across different categories of continual learning methods. First, we mainly compare to CassLe~\cite{gomez2022continually} and LUMP~\cite{madaan2021representational}, which are the state-of-the-art self-supervised continual learning methods. Note that, the reported results of CassLe are reproduced by using the same setup as LUMP for a fair comparison. Second, following~\cite{madaan2021representational}, we also report several self-supervised variants of SCL methods. Specifically, FINETUNE is a vanilla supervised learning method trained on a sequence of tasks without regularization or episodic memory and MULTITASK optimizes the model on complete data. In addition, we also compare to prior SCL methods on self-supervised learning setting. Specifically, we compare against SI~\cite{zenke2017continual} for regularization-based CL methods, PNN~\cite{rusu2016progressive} for architecture-based methods, and DER~\cite{buzzega2020dark} for memory replay method which adapts knowledge distillation by memory replay to match the network logits sampled through the optimization trajectory during continual learning.

\textbf{Dataset.} We compare the performance of SSCL on various continual learning benchmarks using single-head ResNet-18~\cite{he2016deep} architecture. In Split CIFAR-10~\cite{krizhevsky2009learning}, each task includes two random classes out of the ten classes. In Split CIFAR-100~\cite{krizhevsky2009learning}, each task consists of five random classes out of the 100 classes. Split Tiny-ImageNet is a variant of the ImageNet dataset~\cite{deng2009imagenet}, where each task contains five random classes out of the 100 classes with the images sized to 64 × 64 pixels. In Split ImageNet-100, each task consists of 20 random classes out of 100 classes. Here ImageNet-100 is a subset of the ILSVRC2012
dataset with $\approx$ 130k images in high resolution (resized to
224x224).

\textbf{Experimental setting}.  We follow the training and evaluation setup of~\cite{madaan2021representational} for all the SSCL representation learning strategies on Split CIFAR-10, CIFAR-100, and Split Tiny-ImageNet dataset. All the learned representations are evaluated with KNN classifier~\cite{wu2018unsupervised} across three independent runs. In practice, we train all the SSCL methods for 200 epochs and evaluate with the KNN classifier~\cite{wu2018unsupervised}. The memory buffer size is 256 and the models are optimized with SGD optimizer with 0.03 base learning rate under 256 batch size for all the experiments. In addition, following \cite{fini2022self} on the ImageNet-100 dataset, we use LARS optimizer for model training and linear evaluation for testing. 

\textbf{Metrics.}
Following \cite{madaan2021representational}, two metrics are used to evaluate the performance: \textit{Accuracy}, the average final accuracy over all tasks, and \textit{Forgetting}, which measures the forgetting of each task between its maximum accuracy and accuracy at the completion of training. Accuracy and Forgetting are defined as:
{\small
\begin{gather}
    \label{eq:metric}
    Accuracy = \frac{1}{T}\sum\nolimits_{i=1}^{T} A_{T,i} \\
    Forgetting = \frac{1}{T-1}\sum\nolimits_{i=1}^{T-1} \mbox{max}_{t}( A_{T,i} - A_{i,i})
\end{gather}
}%
where $T$ is the number of tasks, $A_{T,i}$ is the accuracy of the model on $i$-th task after learning the $T$-th task sequentially.
Furthermore, we utilize three metrics to measure training efficiency: Training time, we report the training time ratio compared to LUMP baseline which is measured on NVIDIA RTX A4000 GPU; memory, which includes model parameter size, training activation storage, and memory replay buffer size; Flops, which calculate the number of computational operations during backward.

\subsection{Main Results}
As shown in \cref{tab:main_results} and \cref{tab:effic_results}, we evaluate the performance of various SSCL methods,  by using SimSiam~\cite{chen2020simple} and BarlowTwin~\cite{zbontar2021barlow} SSL frameworks  on Split CIFAR-10, Split CIFAR-100 and Split Tiny-ImageNet, respectively. Note that, The training memory cost mainly includes model parameters, memory replay data, and activations of each layer for backward propagation. For SimSiam framework, our method achieves 12\%, 14\%, and 12\% training time reduction, 23\%, 26\%, and 24\% memory, and 33\%, 33\% and 32\% backward FLOPs reduction on three datasets respectively. 
Similarly, for the BarlowTwin framework, our method achieves 13\%, 12\%, and 12\% training time reduction, and 22\%, 21\%, and 22\% memory reduction, and 35\%, 34\% and 33\% backward FLOPs reduction on three datasets respectively.  
Importantly, in terms of forgetting, \textit{our method clearly mitigates the forgetting issue in comparison to all prior methods.} For example, compared to LUMP on Split CIFAR-100 and Split Tiny-ImageNet for Simsiam, we reduce the forgetting by 1.31\%, 1.98\% and 1.21\% respectively with similar accuracy. 

Moreover, we further conduct experiments on the more challenging ImageNet-100 dataset. Specifically, following the same setting with Cassle, we adapted the proposed method on BarlowTwin, and MoCoV2+ respectively. As shown in \cref{tab:image-100}, similar results can be found that the proposed method shows clear gain to improve training efficiency and mitigate catastrophic forgetting with the almost same accuracy. 

\begin{table}[h]
\caption{Accuracy, forgetting, training time, and training memory cost of the learned representations on ImageNet-100 with linear evaluation by using BarlowTwin and MoCoV2 respectively. }
\centering
\scalebox{0.8}{
\begin{tabular}{cccccc}
\toprule
\multicolumn{1}{l}{} & \multirow{2}{*}{Setting} & \multicolumn{4}{c}{ImageNet-100}                                             \\ \cline{3-6} 
\multicolumn{1}{l}{} &                          & Accuracy & Forgeting & \multicolumn{1}{l}{Time} & \multicolumn{1}{l}{Memory} \\ \midrule
\multirow{2}{*}{BarlowTwin} & CassLe & 68.2 & 1.3 & 1x    & 1x    \\
                            & Ours   & \textbf{68.5} & \textbf{0.7} & \textbf{0.88x} & \textbf{0.78x} \\ \midrule
\multirow{2}{*}{MoCoV2}       & CassLe & \textbf{68.0} & 2.2 & 1x    & 1x    \\
                            & Ours & 67.9 & \textbf{1.4} & \textbf{0.89x}  & \textbf{0.79x}   \\ \bottomrule
\end{tabular}}
\label{tab:image-100}
\end{table}

\begin{figure*}[t]
    \centering
    \includegraphics[width=0.24\linewidth]{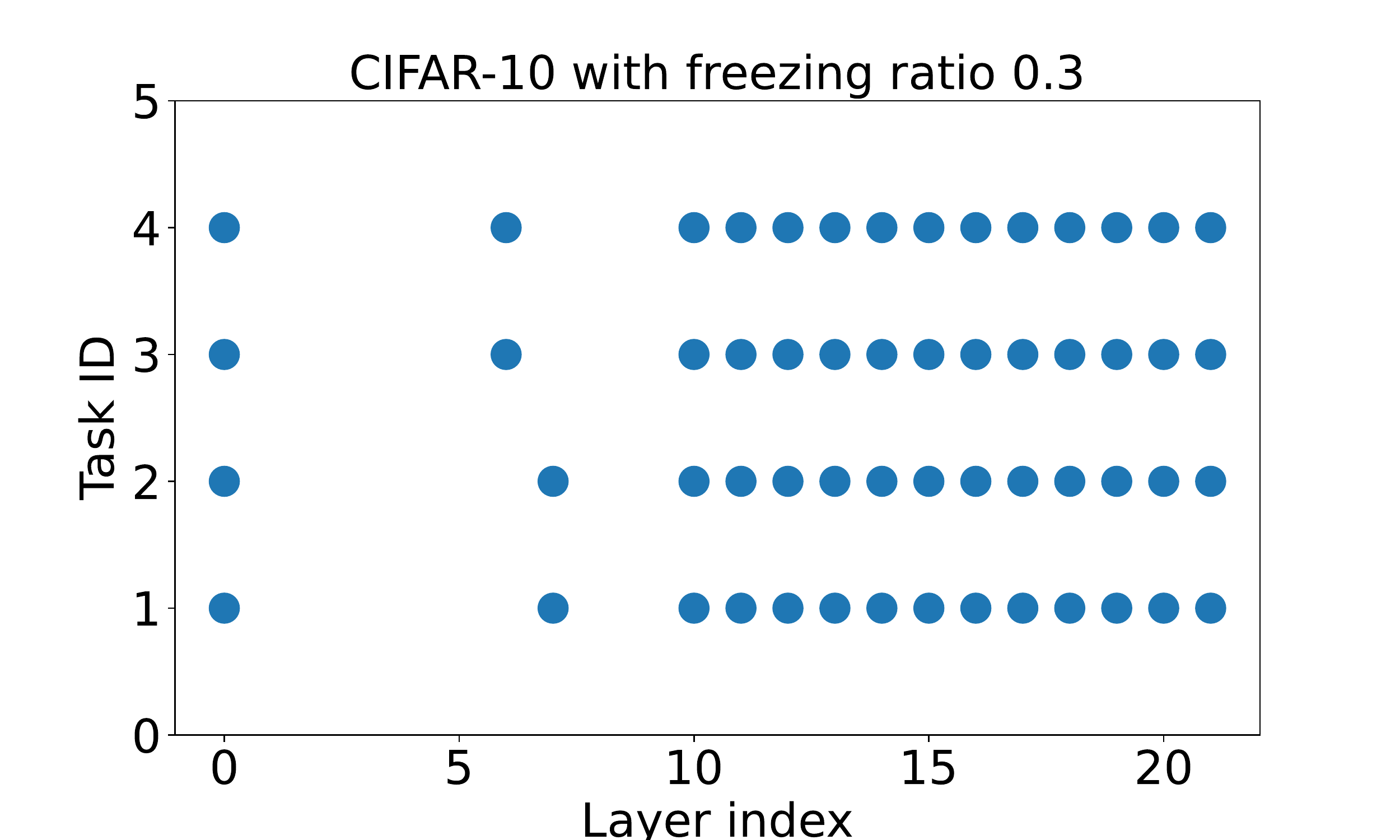}
    \includegraphics[width=0.24\linewidth]{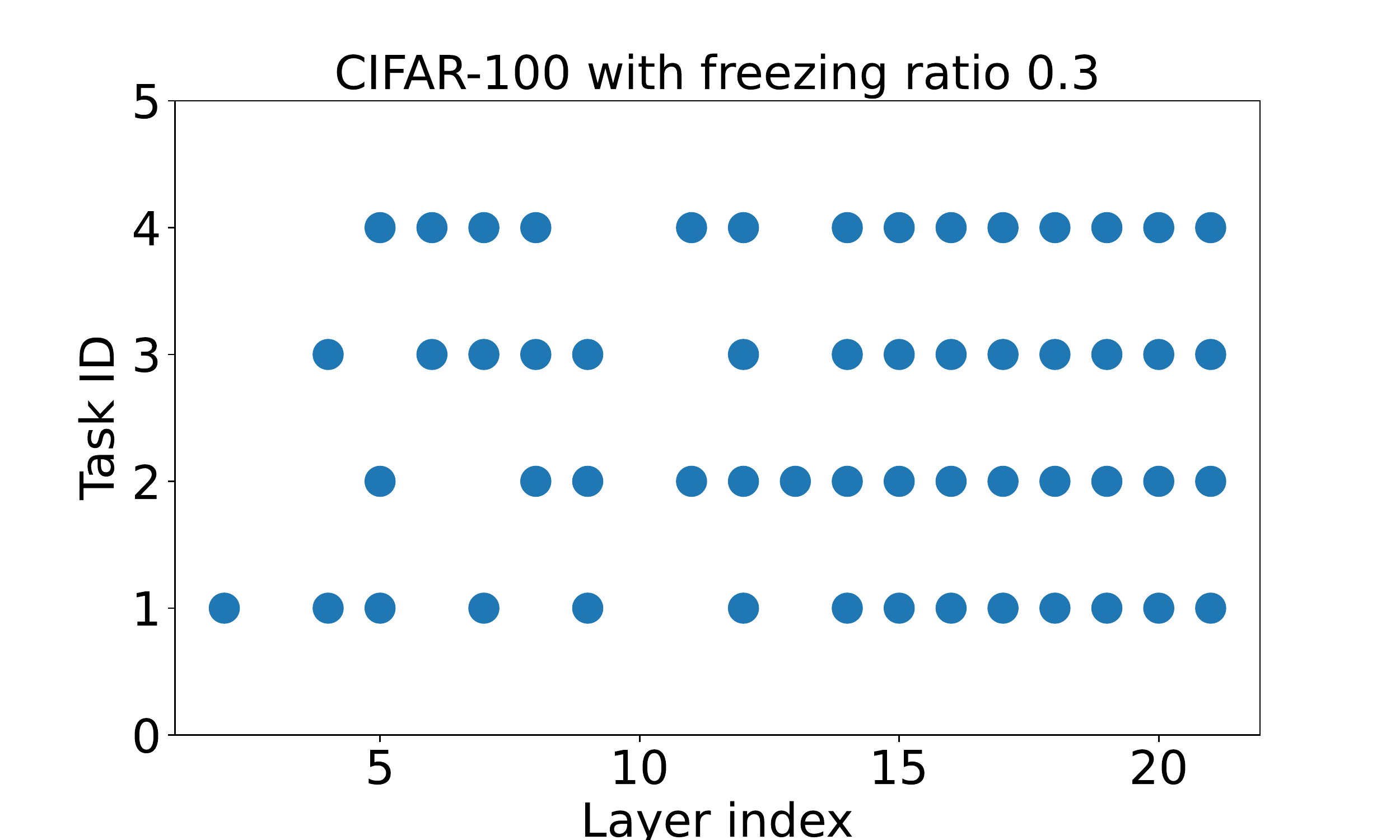} 
    \includegraphics[width=0.24\linewidth]{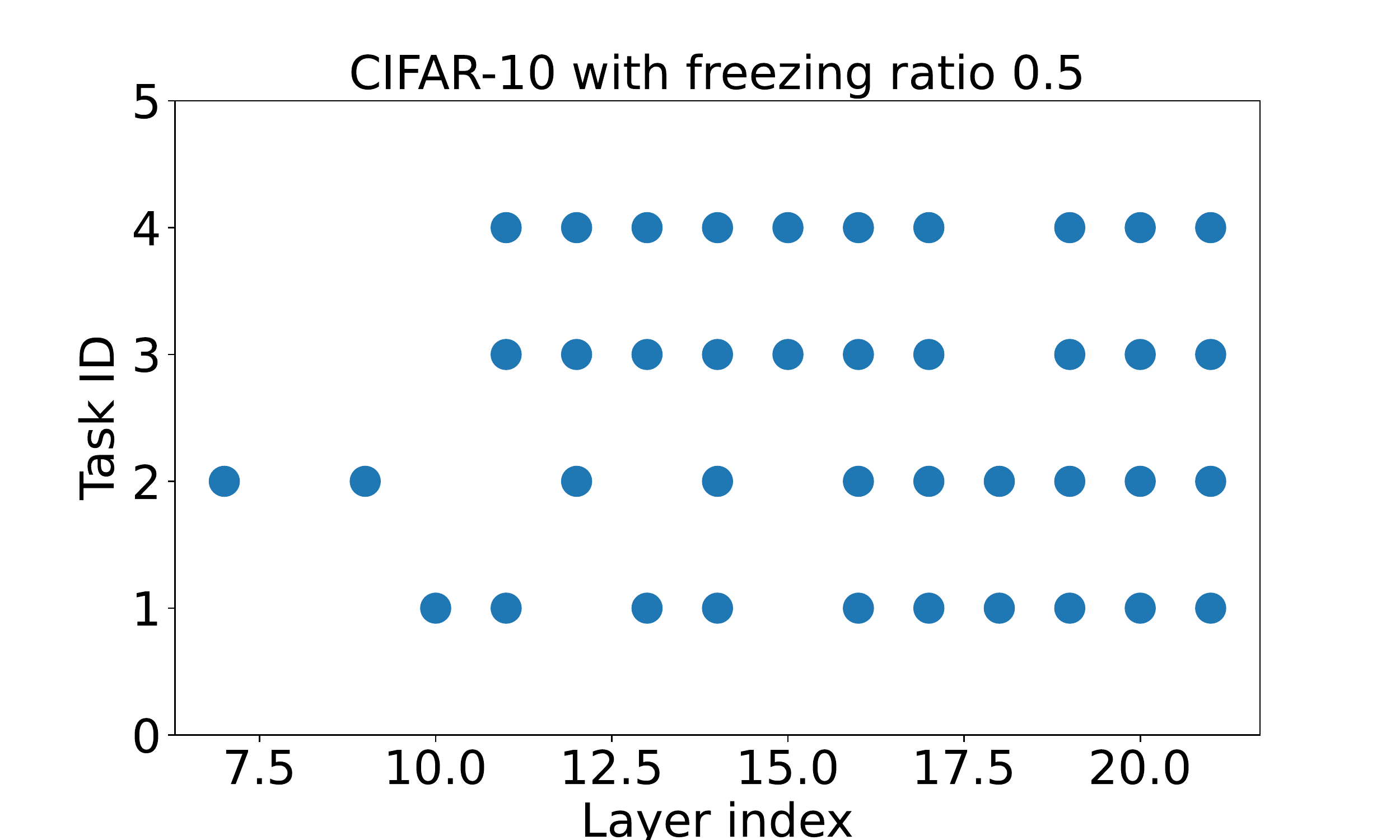}
    \includegraphics[width=0.24\linewidth]{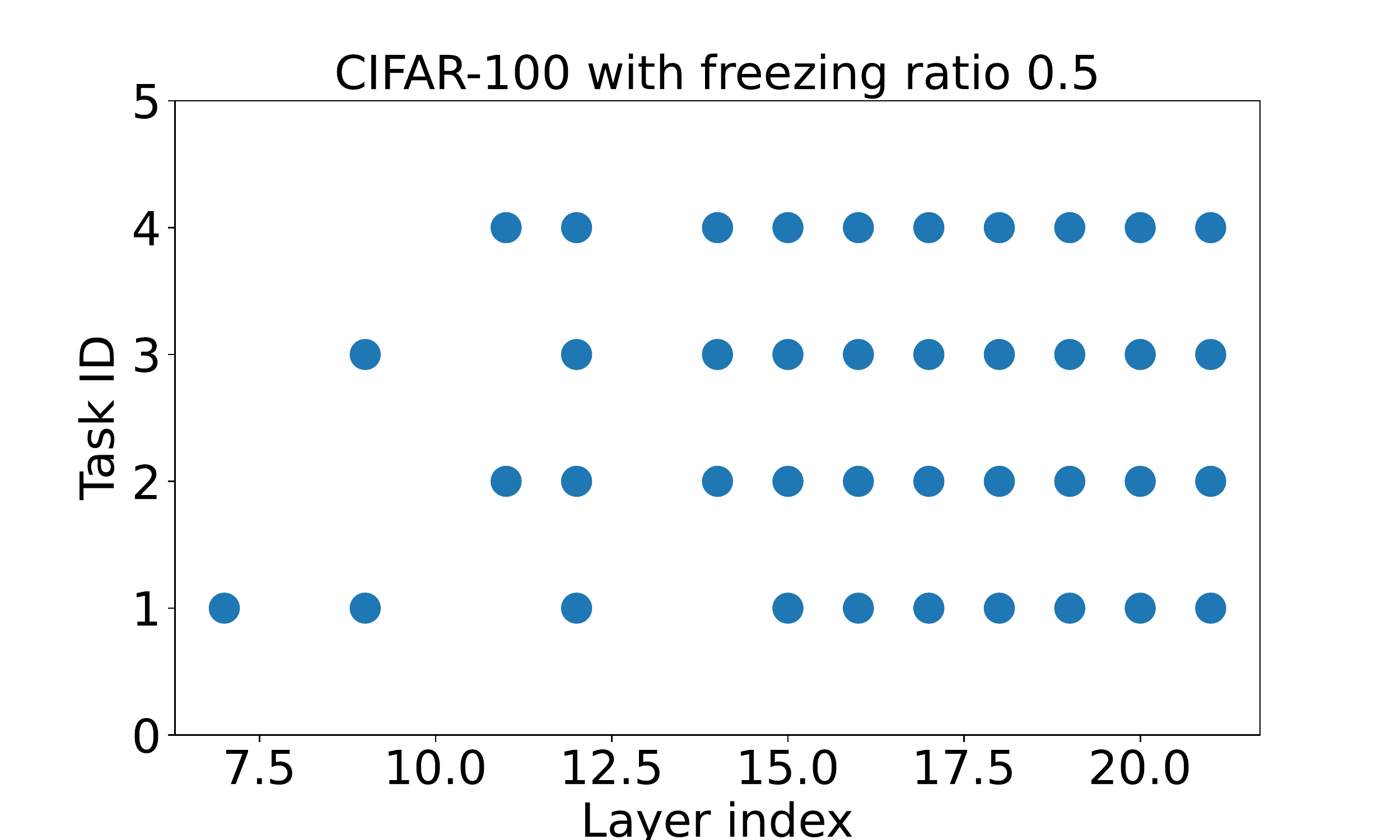}
    \caption{The final selection of updated layer for each task.  The freezing ratios are 0.3 and 0.5 respectively. Note that, each blue point means the index of the updated layer. For Split CIFAR-100 20 tasks setup, we show the first five tasks for simplification.}
\label{fig:layer_dec}
% \vspace{-1em}
\end{figure*}

\vspace{-1em}
\section{Ablation Study and Anaysis}
\label{sec:abla}

\begin{table}[h]
\caption{The ablation study on the proposed method in comparison to layer freezing in ascending layer index (i.e., ``Top layer") order on both Split CIFAR-10 and CIFAR-100 datasets by using BarlowTwin as backbone method.}
\centering
\scalebox{0.9}{
\begin{tabular}{ccccc}
\toprule
\multirow{2}{*}{Setting} & \multicolumn{2}{c}{Split CIFAR-10} & \multicolumn{2}{c}{Split CIFAR-100} \\ \cline{2-5} 
            & Forgetting  & Accuracy & Forgetting & Accuracy\\ \cline{2-5} 
Top layers.        &     0.96 &   89.24         &     {2.57}  &   78.87     \\
Ours        &       \textbf{0.92}   &   \textbf{90.03}       &     \textbf{2.24} &    \textbf{80.54}       \\ \midrule
\end{tabular}}
\label{tab:top_layers}
\end{table}

\paragraph{Task-correlated layer freezing \textit{VS} ascending order layer freezing}
To evaluate the effectiveness of the proposed progressive layer freezing via task correlation, we compare it to ascending order layer freezing which is commonly used in supervised learning settings to improve the training efficiency for a single task. Specifically, for a fair comparison, we also adopt the same cosine annealing to progressively freeze layers under the same freezing ratio (i.e., 0.4).
As shown in the \cref{tab:top_layers}, our method can consistently achieve better accuracy with similar forgetting compared to Naive Selection. The results demonstrate that the task correlation between tasks needs to be considered in SSCL.

\paragraph{The effectiveness of the layer freezing ratio}
To understand the impact of the layer-wise weight freezing ratio, we evaluate the learning performance for four different values of $r$ (i.e., 0.7, 0.6, 0.5, 0.4) on SPLIT CIFAR-10 dataset by using BarlowTwin~\cite{zbontar2021barlow} SCL method as shown in \cref{tab:freeze_ratio}. It can be seen that with the freezing ratio increasing, the learning performance in terms of both accuracy and forgetting gradually degrades, but to a slight extent. As a benefit, the computational costs in terms of both training time and memory significantly decrease.
It makes sense that by fixing more layers, it has a certain degree of negative effect on learning each new task, meanwhile, it indicates less forgetting.
\begin{table}[h]
\centering
\caption{The ablation study on different layer-wise weight freezing ratios. Note that, the ``freeze ratio" means the ratio of the layer numbers to be frozen during the training.}
\scalebox{0.86}{
\begin{tabular}{ccccc}
\toprule
Freeze ratio & Accuracy & Forgetting & Time & Memory \\ \midrule
0.4         &  89.73    &  0.86       &   0.87x         &  0.78x               \\
0.5         &   89.01    &  0.75         &   0.83x          &    0.70x           \\
0.6         &    88.40   &   0.66  &  0.76x          &      0.64x           \\
0.7         &    87.62  &  0.60     &   0.71x      &   0.57x             \\ \bottomrule
\end{tabular}}
\label{tab:freeze_ratio}
\end{table}

% \subsection{The effectiveness of each technique component}
% We study the effectiveness of each technique component in the proposed method on SPLIT CIFAR-10 by using BarlowTwin. As shown in \cref{tab:component}, we consider LUMP as our baseline since we also adapt the Mixup technique to mix the data of current each new task and the old data from the memory replay buffer. First, it can be seen that compared to LUMP, TopK achieves similar accuracy and forgetting but reduces the training time and memory cost by 13\% and 25\%. Second, benefiting from the regularization method, we could further reduce the forgetting by 0.25\% in comparison to LUMP~\cite{madaan2021representational}. 

% \begin{table}[h]
% \centering
% \caption{The ablation study of each technique component. Note that, ``LWF" denotes the proposed large-wise weight freezing method with the ratio as 0.3; ``WPR" denotes that we add the regularization term as shown in \cref{eqt:loss}.}
% \scalebox{0.86}{
% \begin{tabular}{ccccc}
% \toprule
% Method   & Accuracy & Forgetting &  Time &  Memory \\ \midrule
% LUMP  &   89.72       &     1.13       & 1x               &         1x        \\
% + LWF   &    89.85      &        1.21    &    0.87x          &  0.75x               \\
% + WPR    &     89.72     &       0.88     &  0.87x               &        0.75x         \\ \bottomrule
% \end{tabular}}
% \label{tab:component}
% \vspace{-1em}
% \end{table}

% \subsection{Analysis}
% \vspace{-2em}
\paragraph{Progressive task-correlated freezing in SCL}
\label{sec:granul}
% The main reason for using weight freezing in layer-wise is that it can achieve actual training speedup on the general hardware platforms (e.g, CPU and GPU). In addition, due to the high dimension of the model parameters, there also have other fine-grained weight freezing choices, for example, channel-wise freezing which freezes part of weight channels. To evaluate the impact of the weight-freezing granularity, we conduct an ablation study on channel-wise alternative as shown in \cref{tab:granularity}. It is interesting to see that channel-wise freezing has similar  accuracy and slightly worse forgetting in the SSCL setting. 
We evaluate the layer-wise weight freezing on SCL as shown in \cref{tab:granularity}. It is clear to see that layer-wise freezing has obvious performance degradation in both accuracy and forgetting compared to finetune.
This observation further corroborates that the representation learned by SSCL is more general and robust to mitigate catastrophic forgetting in comparison to SCL, which makes it possible to use an aggressive freezing strategy to reduce the computational cost in SSCL.

% \vspace{-1em}
\begin{table}[h]
\scriptsize
\centering
\caption{The effectiveness of layer-wise freezing in both SCL and SCCL settings. Note that, the layer freezing ratio is 0.4. }
\begin{tabular}{cccccc}
\toprule
Setting               & Method       & Accuracy & Forgetting &  Time &  Memory \\ \midrule
\multirow{2}{*}{SCL}                   & Finetune   &     82.87 & 14.26          &      1x         &       1x          \\ 
                  & Layer-wise   &    80.45
      &      16.52      &    0.85x           &  0.72x                \\ \midrule
\multirow{2}{*}{SSCL} & LUMP   &    89.72  & 1.13     &         1x   &         1x      \\
                      & Layer-wise &     \textbf{89.73}     &    \textbf{0.86 }       &    \textbf{0.87x}           &    \textbf{0.75x}             \\
                      % & Ch-wise & 89.72         &      1.05     &         1x      &   1x              \\
\bottomrule
\end{tabular}
\label{tab:granularity}
\end{table}

% \begin{figure*}[h]
%     \centering
%     \includegraphics[width=0.43\linewidth]{cvpr2023-author_kit-v1_1-1/latex/figures/layer_sel.png}
%     \includegraphics[width=0.43\linewidth]{cvpr2023-author_kit-v1_1-1/latex/figures/layer_sel.png}
%     \caption{The selection of updated layer for each task. Note that, each blue point means the index of the updated layer. The freezing ratio is 0.5.}
% \label{fig:layer_dec1}
% \end{figure*}

% \paragraph{Self-supervised continual learning is robust to layer freezing decision}

\vspace{-2em}
\paragraph{Self-supervised continual learning is robust to layer
freezing decision}
We analyze the layer freezing decision by using the freezing ratio of 0.4 for each task on Split CIFAR-10 and Split CIFAR-100 respectively. As shown in \cref{fig:layer_dec}, there are mainly two observations: 1) for inter-tasks, the indexes of layer freezing decisions are highly similar, which means that the selected frozen layers at the first task will not be updated across all the rest tasks. It further helps to show that the learned representation by SSCL is general and robust; 2) for intra-task, it is interesting to see that the first layer and a large number of last layers are updated during training. We conjecture the reason is that the last layers learn high-level features that are sensitive with respect to the given input. In addition, although the first layer learns the general low-level features, the SSCL adapts strong augmentation (e.g., color jittering, converting to grayscale, Gaussian blurring, and solarization) to the input, and should also update the first layer.

% \begin{figure}[h]
%     \centering
%     \includegraphics[width=0.49\linewidth]{figures/corr_t1_rebuttal.pdf} 
%     \includegraphics[width=0.49\linewidth]{figures/corr_t4_rebuttal.pdf}
%     \caption{Layer-wise correlation ratio between two tasks on Split CIFAR-10 5 tasks. Note that, we show the correlation ratios on two consecutive tasks due to limited page space. Other tasks follow the same trend. }
% \label{fig:corr_rebuttal}
% \end{figure}

% \paragraph{\textcolor{blue}{Self-supervised continual learning is more robust compared to supervised continual learning}} 
% \textcolor{blue}{We evaluate the correlation ratios in a fully supervised scenario (SCL) compared SSL case. As shown in Fig.~\ref{fig:corr_rebuttal}, it is interesting to see that 1) both cases have similar correlation ratios in earlier layers; 2) for later layers, the correlation ratios of SSCL are much larger than SCL, which indicates their strong correlations leading to more general representations. }

\section{Conclusion}
In this work, we first investigate the task correlation of SSCL and find that intermediate features are highly correlated between tasks. Based on this, we propose a progressive task-correlated layer freezing method that freezes gradually partial layers with the highest correlation ratios for each task.  Extensive experiments across multiple datasets 
clearly show that our method can significantly improve training computation and memory efficiency meanwhile mitigating catastrophic forgetting,
compared to the SoTA SSCL methods.

% \paragraph{Limitation}  Currently, we conduct experiments on class incremental learning and show that superior performance of training efficiency and forgetting with the same accuracy compared to SOTA SSCL works. In the future, we will further consider other continual learning schemes such as task incremental and domain incremental learning. 

\clearpage

{\small
\bibliographystyle{ieee_fullname}
\bibliography{main_archive}
}

\end{document}

%% file: math_commands.tex
\usepackage{amsmath,amsfonts,bm}

\newcommand{\proj}{\mathrm{Proj}}

\def\eqref#1{equation~\ref{#1}}
\def\1{\bm{1}}

\def\vu{{\bm{u}}}
\def\vv{{\bm{v}}}
\def\vw{{\bm{w}}}
\def\vx{{\bm{x}}}

\def\vy{{\bm{y}}}

\def\mB{{\bm{B}}}

\def\mR{{\bm{R}}}

\def\mU{{\bm{U}}}
\def\mV{{\bm{V}}}

\def\mSigma{{\bm{\Sigma}}}

\DeclareMathAlphabet{\mathsfit}{\encodingdefault}{\sfdefault}{m}{sl}
\SetMathAlphabet{\mathsfit}{bold}{\encodingdefault}{\sfdefault}{bx}{n}

\def\gL{{\mathcal{L}}}

\def\sR{{\mathbb{R}}}

\def\sT{{\mathbb{T}}}